\documentclass[10pt,twocolumn,letterpaper]{article}

\usepackage{iccv}
\usepackage{times}
\usepackage{epsfig}
\usepackage{graphicx}
\usepackage{amsmath}
\usepackage{amssymb}

\usepackage{amsfonts}
\usepackage{booktabs}
\usepackage{multirow}
\usepackage{algorithm}
\usepackage{algorithmic}
\usepackage{subfigure}

\usepackage[breaklinks=true,bookmarks=false]{hyperref}

\iccvfinalcopy % *** Uncomment this line for the final submission

\def\iccvPaperID{****} % *** Enter the ICCV Paper ID here

\ificcvfinal\fi

\begin{document}

%%%%%%%%% TITLE

% \title{Privacy Preserving Semi-Supervised Domain Adaptation via Intra-Domain Mixup}
\title{Uncertainty-Guided Mixup for Semi-Supervised Domain Adaptation without Source Data }

\author{Ning Ma\\
% College of Computer Science, Zhejiang University\\
{\tt\small ma\_ning@zju.edu.cn}
\and
Jiajun Bu\\
% College of Computer Science, Zhejiang University\\
{\tt\small bjj@zju.edu.cn}
\and
Zhen Zhang\\
% College of Computer Science, Zhejiang University\\
{\tt\small zhen\_zhang@zju.edu.cn}
\and
Sheng Zhou\\
% College of Computer Science, Zhejiang University\\
{\tt\small zhousheng\_zju@zju.edu.cn}}

% \and


% \and


% \author{Ning Ma$^{1,2,3}$, Jiajun Bu$^{1,2,3,*}$, Jieyu Yang$^{1,2,3}$, Zhen Zhang$^{1,2,3}$}
% \author{Chengwei Yao$^{1,2,3}$, Zhi Yu$^{1,2,3}$, Sheng Zhou$^{1,2,3}$, Xifeng Yan$^{4}$}
% \thanks{$^*$Corresponding Author}
% \affiliation{%
%   \institution{
%   \textsuperscript{\rm 1}Zhejiang Provincial Key Laboratory of Service Robot, College of Computer Science, Zhejiang University \\
%   \textsuperscript{\rm 2} Alibaba-Zhejiang University Joint Institute of Frontier Technologies \\
%   \textsuperscript{\rm 3} Ningbo Research Institute, Zhejiang University\\
%   \textsuperscript{\rm 4} Computer Science Department, University of California Santa Barbara
%   }
% %   \textsuperscript{\rm 5}Ningbo Research Institute, Zhejiang University}
% }
% \email{{ma\_ning, bjj,yangjieyu,zhen\_zhang,yaochw, yuzhirenzhe,zhousheng\_zju}@zju.edu.cn, xyan@cs.ucsb.edu}

\maketitle
%\ificcvfinal\thispagestyle{empty}\fi

%%%%%%%%% ABSTRACT
\begin{abstract}
   % Domain adaptation methods address domain shift typically by learning domain-invariant features 
   % across available source and target data. However, due to the concerns of privacy or limitations 
   % of communication bandwidth, these methods can be infeasible in practice because of the 
   % unavailability of source data. As a result, source-free domain adaptation methods are 
   % becoming increasingly investigated, with most of which focus on the unsupervised scenario 
   % where the target data is fully unlabeled. Due to the unavailability of target supervision, 
   % these methods could be unstable or fragile in optimization and fail when conditional distribution 
   % shift exists. In this paper, we explore semi-supervised source-free domain adaptation by assuming 
   % a few available labeled target data, which is reasonable in practice. 
   % More specifically, we propose uncertainty-guided Mixup to reduce the 
   % representation's intra-domain discrepancy and perform inter-domain alignment without 
   % directly accessing the source data.
   Present domain adaptation methods usually perform explicit representation alignment by simultaneously accessing the source data and target data. However, the source data are not always available due to the privacy preserving consideration or bandwidth limitation.
   Source-free domain adaptation aims to solve the above problem by performing domain adaptation without accessing the source data. The adaptation paradigm is receiving more and more attention in recent years, and multiple works have been proposed for unsupervised source-free domain adaptation.  However,  without utilizing any supervised signal and source data at the adaptation stage, the optimization of the target model is unstable and fragile. To alleviate the problem, we focus on semi-supervised domain adaptation under source-free setting. More specifically, we propose uncertainty-guided Mixup to reduce the representation's intra-domain discrepancy and perform inter-domain alignment without directly accessing the source data.
%    There are two representation gaps in our proposed setting: 1)the representation gap between the source domain and 
%    target domain; 2) the representation gap between the labeled target data and the unlabeled target data. 
%    To reduce the representation gap between the source domain and target domain, we select source-like examples from
%     the target domain based on uncertainty estimation and explore the vicinal structure of these selected data.
%    To reduce the representation gap between the labeled target data and the unlabeled target data, we adopt a data interpolation 
%    strategy base on Mixup method. Finally, we propose a unified framework consisting of Self-Mixup and Hybrid-Mixup to reduce the two representation gaps.  
   %  
%
We conduct extensive semi-supervised domain adaptation experiments on various datasets.  
Our method outperforms the recent semi-supervised baselines and the unsupervised variant also achieves competitive performance. The experiment codes will be released in the future.
% Experimental results on a number of benchmarks prove the promise of the proposed method by outperforming state-of-the-art unsupervised and semi-supervised approaches.  
\end{abstract}

% Compared to standard mixup method like Mixmatch, the IDM still works well when only given only 
% one or three labeled examples in virtue of the mechanism of source-like selection. 
% to perform intra-group and inter-group data interpolation respectively, where 
   %  the unlabeled target examples with high uncertainty, 
   % the unlabeled target examples with low uncertainty and the labeled target data.
   % The Intra-Domain Mixup consists of Self-Mixup and Hybrid-Mixup, to perform intra-group and inter-group data interpolation respectively. 

%%%%%%%%% BODY TEXT

\begin{figure}[t]
   \begin{center}
   \includegraphics[width=0.99\linewidth]{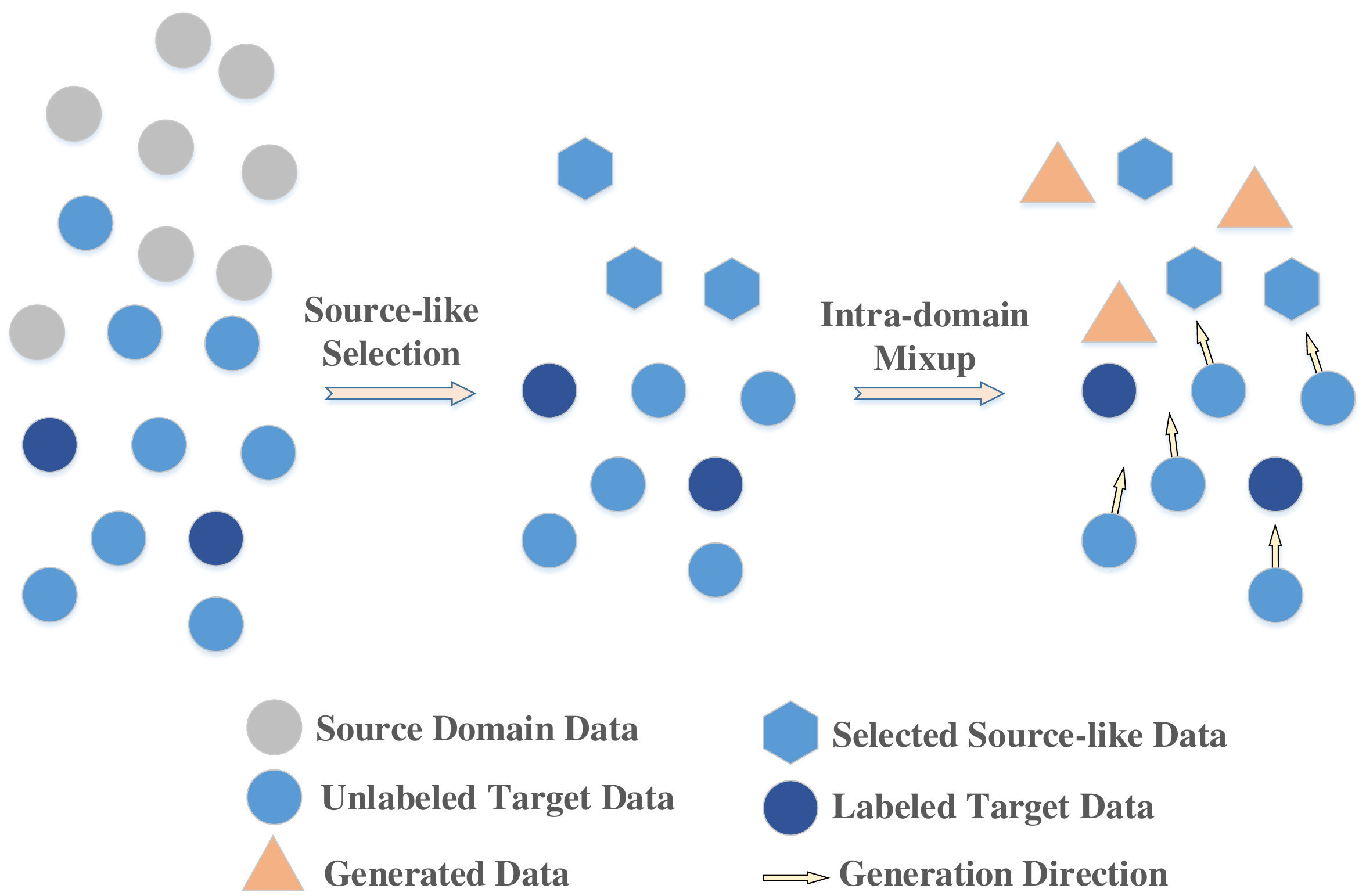}
   \end{center}
      \caption{The illustrations of our motivation. The source domain data are not available at the adaptation stage due to privacy preserving consideration.
      We firstly select the unlabeled target data most similar with the source domain, then generate new examples around 
      the selected data to explore the vicinal structure and bridge the domain gap.
      }
   \label{fig:motivation}
   \end{figure}
   
\section{Introduction}
% Present deep neural networks (DNNs) have got success when dealing with various tasks such as image classification and object detection. 
% Present deep neural networks (DNNs) trained on source data usually suffer
%  performance degradation on target data, due to the target data do not always have the same 
%  distribution with the source data. Hence Domain Adaptation(DA) is proposed and has 
%  been proved to be a powerful technique solving the distribution shift problem. In the common setting of DA, 
%  the source domain has sufficient examples with labels, the target domain 
% has different but related examples with no or few labels provided. Based on the above setting, 
% the dominant DA paradigm stresses explicit 
% representation alignment by adversarial training \cite{GAN} et al. 

Deep neural networks (DNNs) typically suffer significant degradation when the distribution of test data 
is different from training data, which is commonly-encountered in machine learning practice. 
Domain Adaptation (DA) has been shown to be an important technique to address such \emph{distribution shift} or
\emph{domain shift}. In the common setting of DA, 
 the source domain has sufficient examples with labels, the target domain 
has different but related examples with no or few labels provided. Based on the above setting, most methods 
simultaneously access the data from both domains and propose to reduce the domain gap by learning domain-invariant features.
However, due to concerns of 
data privacy or the limitations of communication bandwidth, these methods could be infeasible in practice 
because of the unavailability of source data.

Recently, source-free domain adaptation is receiving increasingly popularity. 
These methods generally follow a pre-train and fine-tune paradigm, \emph{i.e,}, 
pre-training models on the source domain, and performing fine-tuning adaptation 
using target data only. In unsupervised setting, there have been several methods 
performing source-free adaptation on the target domain, such as fixing the pre-trained 
classifier layer with updating the others (SHOT, \cite{SHOT}), updating the Batch 
Normalization layer with fixing the others (Tent, \cite{wang2021tent}), updating 
all layers with augmented target data \cite{Li_2020_CVPR}. 
However, due to the unavailability of any labels for supervision, most of these 
methods depends on predicted pseudo labels, which are inherently unreliable and 
generally lead the model optimization to be unstable and fragile with false feature 
alignments. In this paper, we approach source-free domain adaptation in a 
semi-supervised manner by assuming few labeled target data is additionally available, 
which is quite reasonable in practice and achieves significantly improved performances. 

Semi-Supervised DA (SSDA) aims to boost the adaptation quality with a few labeled target data and has a long research history \cite{NIPS20104009,xiao2012semi,MME}.
% boost the adaptation quality with a few labeled target examples and has a long research history.
To solve the source-free SSDA, two essential problems must be considered.  
The first is how to reduce the representation's intra-domain discrepancy 
between the labeled target data 
and the unlabeled target data. This problem arises because of different training objectives for the 
labeled target data and the unlabeled target data.
To reduce this intra-domain discrepancy, traditional semi-supervised learning methods 
usually need a considerable proportion of labeled examples.  
However, performing source-free SSDA makes a great challenge for these 
traditional techniques. Because in the common setting \cite{MME}, 
only one or three labeled examples per class can be utilized.  
As the basics of our solution, we consider linear interpolation 
between the labeled target examples and the unlabeled target examples, 
making the model exploring the structure between the two parts. 
The linear interpolation methods for deep neural networks were proposed by 
Mixup \cite{zhang2018mixup}, which trains a neural network on convex 
combinations of pairs of examples and their labels. 

% we observe that a part of unlabeled target data have similar representation with the source examples. 
% Furthermore, when meeting large-scale adaptation datasets like DomainNet, the entropy 
% regularization is inferior to balance the large category \cite{feng2021kd3a}. 
% Hence, these unlabeled target examples can be treated as the agents of source examples to perform cross domain alignment. by the combination of Monte Carlo Dropout \cite{mcdropout} on the model and random transformations on data
% Furthermore, we can treat these unlabeled target examples as the agents of source examples to perform cross domain alignment. 

The second problem is how to perform effective domain alignment while not accessing the source domain. 
We use a common assumption that  domain shift makes the target distribution different but related with the source distribution. Inspired by the semi-supervised setting, our motivation is to estimate the related part, in which the unlabeled target examples have a similar representation with the source examples.  To this end, we propose source-like selection based on uncertainty estimation considering the  uncertainty of data and model. The unlabeled target examples with low uncertainty have two merits for the adaptation task: 1) 
the model usually produces high quality pseudo labels for the unlabeled target examples with low uncertainty, and these high quality pseudo labels directly provide a substantial supplement for the labeled target examples.
 2) the unlabeled target examples with low uncertainty are usually similar with source examples in representation space, 
 hence pushing the representation of the  high uncertainty examples toward the low uncertainty examples nearly equals aligning the cross-domain representations.
 
 To solve the above two problems under an unified framework, we also use linear interpolation to reduce the domain gap.  Figure \ref{fig:motivation} illustrates our motivation and proposed solution for domain alignment.  
 % consisting of Self-Mixup and Hybrid-Mixup to reduce the two representation gaps. , 
% when the prediction uncertainty was   For the target examples, the number of low uncertainty examples is usually smaller
% than the rest. 
% measured by Shannon Entropy \cite{entropy}, 
% In section \ref{SourceLikeSelection}, we examples most similar with the source examples.
We conduct extensive semi-supervised domain adaptation experiments on various datasets such as  large scale dataset DomainNet and two small size datasets Office-31, Office-Home . 
The results show that our method outperforms the recent baselines of SSDA. Moreover, our unsupervised version using source-like selection also outperforms the recent unsupervised baselines under source-free setting.
We formulate our method as Uncertainty-guided Intra-Domain Mixup (UIDM). Our contributions are summarized as follows:
\begin{itemize}
\item We summarize the recent domain adaptation methods considering source-free setting, and found that the optimization of target model is easy to corrupt without any supervised signal and source data.
This motivates us to  propose a semi-supervised version under this setting.  

\item For the source-free SSDA, we propose uncertainty-guided mixup by leveraging uncertainty estimation to reduce the intra-domain discrepancy and inter-domain representation gap. 
% \item 
% \item  At the adaptation stage, we split the target examples into three groups: the unlabeled target examples with high uncertainty,  the unlabeled target examples with low uncertainty and the labeled target examples.  The IDM Hybrid-Mixup and Self-Mixup to perform inter-group and intra-group data interpolation.

% The intra-domain mixup consists of two critical operations: Hybrid-Mixup, mixuping the labeled target smaples and the unlabeled target examples, 
% and mixuping the unlabeled target smaples with low uncertainty and the unlabeled target examples with high uncertainty. 

\item We evaluate the performances against different baselines on various domain adaptation datasets.  Compared with the recent SSDA methods using both 
domains for adaptation, our framework outperforms these methods in almost all the tasks while not accessing the source domain at the adaptation stage.  
\end{itemize}

\section{Related Work}  
\subsection{Source-Free Domain Adaptation}
Domain Adaptation aims to deal with distribution shift problem by transferring knowledge 
from a source domain to a different but related target domain. Deep neural networks have been recognized with better
representation ability compared to traditional methods.  Among the recent deep learning techniques, 
Adversarial Deep Learning \cite{GAN} gets strong performance in many adaptation scenarios by minimizing the domain gap with generative models.  We recommend referring to the surveys \cite{zhuang2020comprehensive} for more details.

In the meantime, data privacy is becoming more important as deep neural networks get success in domain adaptation.
There exist various methods aiming at source-free for source data such as Federated Learning \cite{Peng2020Federated}.
However, these methods lack the flexibility to be deployed in practical tasks, because of the requirement to access source data.

Recently, some works such as \cite{kddsourcefree} show the possibility to 
perform adaptation on target domain while not accessing source data. Concretely, \cite{kddsourcefree} proposes source-free 
domain adaptation and set the basic framework of pre-training and finetuning. Based on the framework, 
SHOT \cite{SHOT} adopts the Source Hypothesis Transfer, which fixes the pre-trained classifier and finetunes other modules 
at adaptation stage. \cite{Li_2020_CVPR} propose model adaptation methods by iteratively generating the source-style
 examples and updating the model. Tent \cite{wang2021tent} updates normalization layer and fixes others when performing 
 adaptation. 

\subsection{Semi-Supervised Learning for DA}
Semi-Supervised Learning aims to use the labeled and the unlabeled data together 
to facilitate the overall learning performance. 
 Typical methods can be classified into three categories: 
 i) consistency regularization, encourage the model to have stable outputs when applying perturbations to examples \cite{tarvainen2017mean,VAT}.
 ii) entropy minimization, encouraging the model to make a confident prediction for unlabeled examples \cite{EM}
 iii) data augmentation, by performing linear interpolation \cite{zhang2018mixup,MixMatch} 
 or self-supervised learning with more data transformations \cite{mishra2021surprisingly}.
In the traditional semi-supervised setting, where a considerable amount of labeled examples are available. However, there are only one or three labeled target examples in the setting of Semi-supervised Domain Adaptation (SSDA) \cite{MME}.    
% To solve this chanlange, It has a long research history for the 
% SSDA problem \cite{NIPS20104009,xiao2012semi,Donahue2013CVPR,yao2015semi,AAAI1714538,DingTIP,8444719,MME,kim2020attract,li2020online,qin2020opposite,BiAT}. 
To solve domain shift challenge in semi-supervised setting, \cite{NIPS20104009} proposed a co-regulation based approach for SSDA, 
based on the notion of augmented space;
\cite{xiao2012semi} proposed a kernel matching method mapping the labeled source data to the target data.
Recently, MME \cite{MME} proposed minimax entropy method and works well on DomainNet \cite{peng2018moment}, which is used as a large domain adaptation dataset. Furthermore, 
many recent works follow the setting of MME and get more strong performance, e.g., \cite{BiAT,APE}. Among the published methods, none of them solves the problem under source 
free setting.
  
\subsection{Uncertainty Estimation}
Estimating the noise from input data or models is important for the deep learning community.
Among existing estimation methods, Bayesian networks\cite{blundell15} are typical tools to predict the uncertainty of weights in the network. 
Using Bayesian theory, \cite{wtat} attempt to provide prediction results with confidence.
\cite{rizve2021in} propose an uncertainty-aware pseudo-label selection framework to improve pseudo labeling accuracy. 
In domain adaptation tasks, \cite{UEDA} get strong performance on the domain adaptive semantic segmentation with uncertainty. Unlike the previous works, we only consider the 
prediction uncertainty on the target domain using entropy metric, which means lower prediction entropy indicates lower uncertainty. Further, we consider not only the data uncertainty
brought by random transformation and linear interpolation, but also the model uncertainty brought by the Dropout operation.

\begin{figure*}[t]
   \begin{center}
   \includegraphics[width=0.80\linewidth]{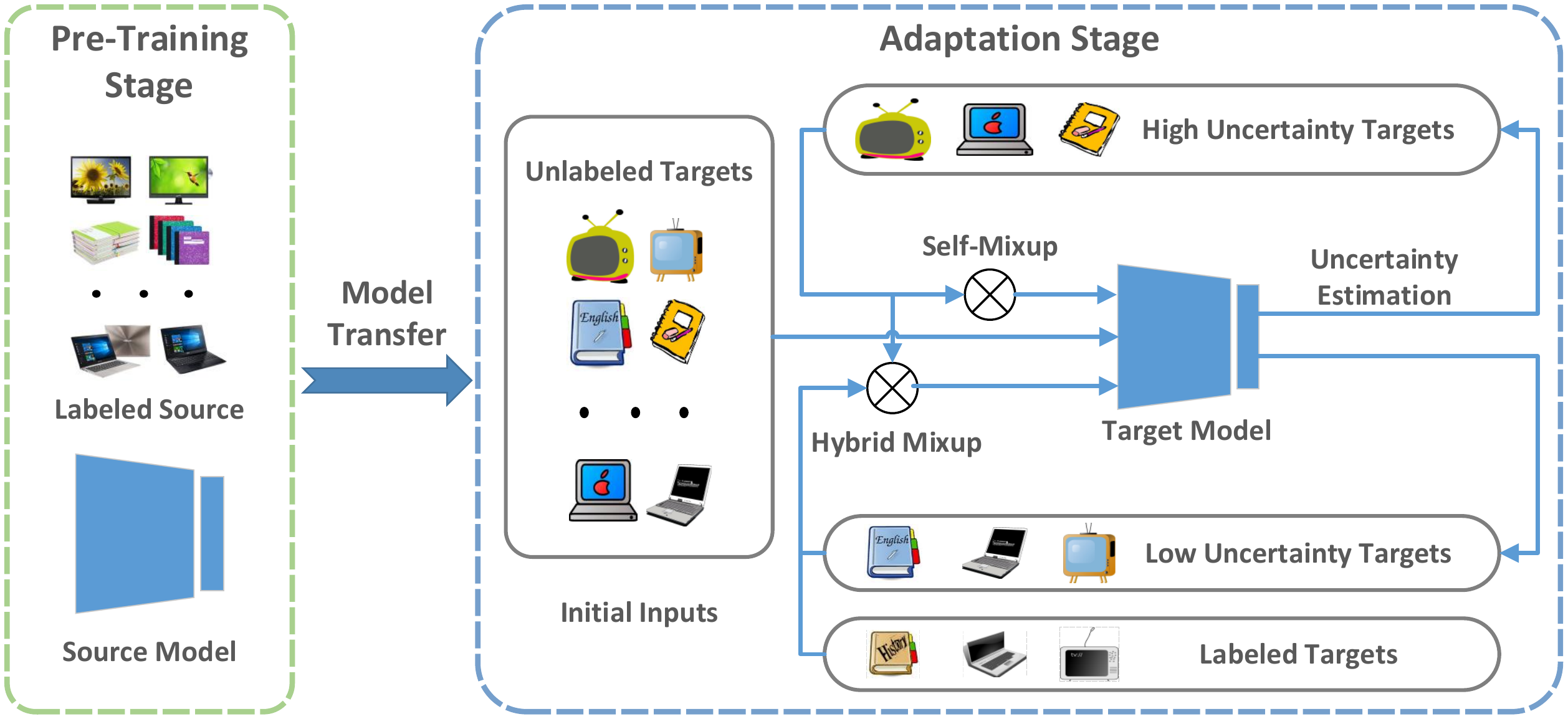}
   \end{center}
      \caption{The overall framework of our methods consisting of the pre-training stage and the adaptation stage.
       At the adaptation stage, we split the target examples into three groups: 
       the unlabeled target examples with high uncertainty, 
       the unlabeled target examples with low uncertainty 
       and the labeled target data. Then we use Hybrid-Mixup and Self-Mixup to perform inter-group and intra-group data interpolation. (The data flows for Self-Mixup are simplified for clarity.)
       Finally, the interpolated data was utilized to update the encoder of the target model.
      }
   \label{fig:framework}
   \end{figure*}

\section{Proposed Framework}

\subsection{Problem Formulation}
Here we introduce the adaptation process of source-free DA. On the source domain,
we have labeled data $\mathbf{X}^{s} \in \mathbb{R}^{n\times d} $ and corresponding labels $\mathbf{Y}^{s} \in \mathbb{R}^{n \times 1}$.
On the target domain, we have a few labeled data $\mathbf{X}_{l}^{t} \in \mathbb{R}^{o \times d} $ 
with corresponding labels $\mathbf{Y}_{l}^{t} \in \mathbb{R}^{k \times 1} $, 
and the unlabeled data $\mathbf{X}_{u}^{t} \in \mathbb{R}^{m\times d}$. 
Here the superscripts $s$  and $t$ denote the source domain and the target, respectively.
 $n$ , $o$ and $m$ denote the number of the source data,
  the labeled target data and the unlabeled target data, respectively. 
The universal setting of DA is training a model to predict the labels of  $\mathbf{X}_{u}^{t} $, 
with the help of $\mathbf{X}^{s}$ and $\mathbf{Y}^{s}$. 
Inspired by the framework of HTL, we firstly define a encoder function $f(x): \mathbf{R}^{d} \rightarrow  \mathbf{R}^{h}$  
and an objective function $L_{w}(x,y)$  with learnable parameters 
$ \mathbf{W} \in \mathbb{R}^{k \times h}$. 

On the source domain, we optimize function $f$ and $ \mathbf{W}$ by :
\begin{equation}
   \begin{array}{ll}
   L_{w}\left(\mathbf{X}^{s}, \mathbf{Y}^{s}\right)= \\
   -\mathbb{E}_{\left(x, y\right) \sim \mathbf{X}^{s},\mathbf{Y}^{s} } 
   \sum_{k=1}^{K} y_{k} \log S_{k} \left( \mathbf{W} \times f\left(x \right) \right),
   \end{array}
   \end{equation}
where $K$ is the class number, $S(·)$ is Softmax function. Note that in the origin work of HTL \cite{SHOT}, $\mathbf{W}$ is the parameters of classifier consisting of one fully connected layer. We regard $\mathbf{W}$ as a part of 
loss function $L_w$ for the sake of analysis. 

On the target domain, we fix $\mathbf{\hat{W}}$ that has captured the statistics of the source data, and update the encoder with the proposed UIDM method. The next contents are organized as follows:
  we firstly introduce the  mixup operation and its extension applied on the target domain. Then we analyze what the model can transfer from the source to the target and design the source-like selection methods using uncertainty estimation. Finally, we 
put all the processes into an unified framework including Self-Mixup and Hybrid-Mixup. Figure \ref{fig:framework} demonstrates the proposed framework. 
% On the target domain, we focus on the following problems:  to reduce the intra-domain discrepancy between the labeled target examples and the unlabeled target examples;  to perform efficient representation alignment while not accessing the source domain. 

% 1.how to reduce the representation gap between the labeled target data and the unlabeled target data.
% 2.how to perform efficient representation alignment while not accessing the source domain?
% for problem 1, we adopt mixup methods, which performance linear interpolation 
% between the labeled target data and the unlabeled target data. For problem 2, we observe 
% that a part of unlabeled target data have a similar representation with the source data. 
% Hence, these labeled target data can be treated as the source data to perform cross domain alignment.

\subsection{Target Mixup}
 To reduce the intra-domain discrepancy between the labeled target examples and the unlabeled target examples, we try to use Mixup \cite{zhang2018mixup} to explore the structure of the two parts. Mixup aims to train a neural network on convex combinations of pairs of examples and their labels. The simple method has been shown powerful ability to improve robustness to adversarial examples,  and reduce corruption risk for training generative adversarial networks \cite{goodfellow2014generative}. 
 Therefore in earlier works \cite{Xu_Zhang_Ni_Li_Wang_Tian_Zhang_2020}, domain mixup is used to boost Adversarial Domain Adaptation and get satisfactory performances. 
 However, the Adversarial Domain Adaptation needs source  domain to be accessable, which conflicts with our source-free setting.
 In our setting, we only consider mixup operation on the target domain. The new example $\{ \mathbf{x}, \mathbf{y}\}$ generated by mixup $M\left((\mathbf{X}_1,\mathbf{Y}_1), (\mathbf{X}_2, \mathbf{Y}_2)\right)$ can be defined as:
%  the mixup sample $\thicksim {x}$ and corresponding label can be combined as:
%  $\hat{Y}_{u}^{t}$ denotes the predicted classes for ${X}_{u}^{t}$.${X}_{l}^{t}$ and ${X}_{u}^{t}$. 
% As implementation,
 \begin{equation}
   \begin{array}{ll}
    \mathbf{x}= \lambda\mathbf{ x}_{1} + (1-\lambda) \mathbf{x}_{2} ;  \mathbf{x}_{1} \in \mathbf{X}_1, \mathbf{x}_{2} \in \mathbf{X}_2\\
    \mathbf{y}= \lambda \mathbf{y}_{1} + (1-\lambda) \mathbf{y}_{2} ;  \mathbf{y}_{1} \in \mathbf{Y}_1, \mathbf{y}_{2} \in \mathbf{Y}_2
   \end{array},
   \label{MMM}
\end{equation}

where $\lambda$ denotes the mixup coefficient,
For implementation, we adopt MixMatch \cite{MixMatch}, 
a improved version of mixup which performs linear interpolation by 
$M((\mathbf{X}_{l}^{t}, \mathbf{Y}_{l}^{t}), (\mathbf{X}_{a}, \mathbf{\hat{Y}}_{a})$ and  $M((\mathbf{X}_{u}^{t},  \mathbf{\hat{Y}}_{u}), (\mathbf{X}_{a}, \mathbf{\hat{Y}_a})) $
where $\mathbf{X}_{a}$ is the union of augmented $\mathbf{X}_{l}^{t}$ and $\mathbf{X}_{u}^{t}$,  
$\mathbf{\hat{Y}}_{a}$ is the label set corresponding to $\mathbf{X}_{a}$,
$\mathbf{\hat{Y}}_{u}$ is the soft labels of $\mathbf{X}_{u}^{t}$.
However,  the original Mixmatch above can not cope with our 
situation well. In our setting, only one labeled target examples making 
$M((\mathbf{X}_{l}^{t}, \mathbf{Y}_{l}^{t}), ( \mathbf{X}_{a}, \mathbf{\hat{Y}}_{a})$ often lost its function.  This impels us to select some 
source-like examples from $\mathbf{X}_{u}^{t}$ as the labeled data. In the following section, we will define 
what is the source-like examples and design selection methods by uncertainty estimation.

\subsection{Source-Like Selection}
\label{SourceLikeSelection}
To produce high quality pseudo labels and push the target representation  towards  the source representation, we want to select the unlabeled targets which  have similar representation with source examples as the agents of source domain.
Before the introducing for selection operation, we analyze what the model can transfer from the source domain to the target domain.
Let $\mathbf{W}$ denote the parameter matrix of the model's fully connected classifier. Though the training 
on the source domain, the estimated matrix $\mathbf{\hat{W}}$ can capture the expected representations 
for each class when the model is optimized. If $ \mathbf{\hat{W}}_{k}$ denote the $k$-th vector corresponding to $k$-th class, 
 $f(x)$ denotes the representation of input $x$,  the following equation will be hold when the model is fully optimized:

\begin{equation}
   \mathbf{\hat{W}}_{k}= \mathbb{E}_{\mathbf{x}_{k} \in \mathcal{X}^{s}}[f(\mathbf{x}_{k})],
\end{equation}

where $\mathbf{x}_{k}$ denote the example with class $k$.  If we transfer $\mathbf{\hat{W}}$ to the target domain, we can get the following objective function:
\begin{equation}
   \begin{array}{ll}
   \mathcal{L}_{cos}\left( \mathbf{X}^{t}, \mathbf{Y}^{t}\right)= \\
   -\mathbb{E}_{\left(x, y\right) \sim \mathbf{X}^{t},\mathbf{Y}^{t} } 
   \sum_{k=1}^{K} y_{k} \log S_{k} \left(  \mathbf{\hat{W} }\times f\left(x \right) \right),
   \end{array}
   \label{src_objective}
   \end{equation}

Due to $\mathbf{\hat{W}}$ is fixed when transferred to the target domain. 
Equation \ref{src_objective} indicates we can get lower $\mathcal{L}_{cos}\left(\mathbf{X}^{t}, \mathbf{Y}^{t}\right)$ 
when the representation of an unlabeled target example similar
to a vector of $\hat{W}$, i.e., the example is close to a class of source data in representation space.
This gives us the motivation to explore and exploit the unlabeled target examples 
which have high confidence close to the source data in representation space.

% \subsubsection{uncertainty based target exmple selection}
Next, we introduce a simple method to select high confidence target examples based 
on uncertainty estimation.  If we let $p(x)=S(\mathbf{\hat{W}} \times f(x) )$ denote the class probability vector, we can use the  
entropy $ e=\sum_{k=1}^{K} P_{k} log P_{k}$ to estimate the uncertainty of sample $x$, 
where $P$  is derived from:
\begin{equation}
   {P}= 1/2\times n_{r} \times \sum_{i=1}^{n_{r}} D_{i}(p(x_1)+p(x_2))
   \label{cal_P}
\end{equation}
Note that we use random transformation to consider the uncertainty from data, i.e., $x_1$ and $x_2$ are two different random transformation examples from input sample $x$. 
% Similar works such as SimCLR[], in which multiple transformations are integrated into its 
% framework to perform contrastive learning.
Furthermore, to consider the uncertainty brought from the model, 
we repeat $n_{r}$ times Dropout operation $D_i$ to calculate $p(x)$ using MC-Dropout \cite{pmlr-v48-gal16} method.
MC-Dropout is an uncertainty estimation method opening Dropout in $f(x)$ and 
calculate the mean of repeated outputs in the estimation stage.  

Finally, we consider how to select the unlabeled target examples from their estimated uncertainty.
Let $\mathbf{I}_{k}$ denote the unlabeled target examples with predicted class $k$. 
For each $k$, we sort  $\mathbf{I_{k}}$ according to their uncertainty with ascending order, and select top-$h$ samples.   Compared with SHOT, the source-like selection can produce high quality pseudo labels for a part of target examples and avoid redundancy noises.  Furthermore, these selected examples can be set as the agents of source domain to perform domain alignment. 

% We leave more efficient selection as a future work.

\subsection{Model Training}

In this section, we provide the overall objective function training  our model.
We split $\mathbf{X}_{u}^{t}$ into the selected samples $\{\mathbf{I}, \mathbf{\hat{Y}}\}$  and 
the rest samples $\{\mathbf{I}_{r},  \mathbf{\hat{Y}}_{r}\}$, where $\mathbf{\hat{Y}}$ and $\mathbf{\hat{Y}_{r}}$ are predicted soft labels.
Derived from MixMatch\cite{MixMatch}, the interpolation of examples can be  re-defined as:

\begin{equation}
   (\mathbf{X},  \mathbf{Y})=M_{hybrid}\left((\mathbf{{X}}_{l}^{t} + \mathbf{I} , \mathbf{Y}_{l}^{t} + \mathbf{\hat{Y}}), (\mathbf{I}_{r},  \mathbf{\hat{Y}}_{r})\right),
   \label{LLL}
\end{equation}

\begin{equation}
   (\mathbf{U},\mathbf{Y}_u) = \bigcup_{i}      M_{self}\left(( \mathbf{\hat{X}}_{i},  \mathbf{\hat{Y}}_{i}), (\mathbf{\hat{X}}_i,  \mathbf{\hat{Y}}_i) \right),
   \label{UUU}
\end{equation}
where $\mathbf{\hat{X}}_i$ includes $\mathbf{{X}}_{l}^{t}$, $\mathbf{I}$ and $\mathbf{I}_{r}$, and $\mathbf{\hat{Y}}_i$ denotes corresponding ground truth labels or soft labels.
Following \cite{MixMatch}, the loss for $(\mathbf{U},  \mathbf{Y}_u)$ can be defined using Mean Squared Error:
\begin{equation}
   \begin{array}{ll}
   \mathcal{L}_{mse}\left(\mathbf {U}, \mathbf{Y}_u \right)= \\
   -\mathbb{E}_{\left(x, y\right) \sim \mathbf{U},\mathbf{Y}_{u}} 
   \sum_{k=1}^{K} \left | y_{k} - S_{k} \left( \mathbf{ \hat{W}} \times f\left(x \right)  \right) \right|^2,
   \end{array}
   \label{src_objective}
\end{equation}

Therefore, we get the mixup loss:  
\begin{equation}
   \mathcal{L}_{m} =\mathcal{L}_{cos}(\mathbf{X}, \mathbf{Y})+ \alpha \mathcal{L}_{mse}(\mathbf{U}, \mathbf{Y}_u),
   \label{LOSS}
\end{equation}
where $\alpha$ is the weight of $\mathcal{L}_{mse}$.  Algorithm \ref{ALG} demonstrates the implementation of the UIDM at the adaptation stage.

\subsection{Generalization Analysis}
In this section, we try to theoretically explain why UIDM works and the differences between UIDM and traditional  mixup methods.  We use  Integral Probability Metric \cite{IPM}  to denote the expected distance between the source domain and the target domain:

\begin{equation}
   \label{proposed_bound}
   \gamma_{\mathcal{H}}(\mathbb{S}, \mathbb{T}):= \sup_{h \in \mathcal{H}   } \left|\mathbb{E}_{\mathbb{S}} h(z)-\mathbb{E}_{\mathbb{T}} h(z)\right|,
   \end{equation}
where $h: \mathcal{X} \times \mathcal{Y} \rightarrow  {R} $ is a real-value function and $\mathbb{S}$ and $\mathbb{T}$ denote the distribution of the
source domain and the target domain respectively. If we only consider binary classification, according to \cite{RRR}, for any $B$-uniformly
bounded and $L$ Lipchitz function $\Gamma$,for all $h \in \mathcal{H}$,with probability at least $1-\delta$:
\begin{equation}
   \label{proposed_bound}
   \gamma_{\mathcal{H}}(\mathbb{S}, \mathbb{T})   \leq  \gamma_{\mathcal{H}}(\hat{\mathbb{S}}, \hat{\mathbb{T}}) + 2L \mathcal{R}(h|\mathbf{Z})+ B \sqrt{\frac{\log (1 / \delta)}{2 n}},
   \end{equation}
where $\mathcal{R}(h|\mathbf{Z})$ is the Rademacher Complexity \cite{RRR} of the function class $\mathcal{H}$ w.r.t the target examples $\mathbf{Z}$, $\hat{\mathbb{S}}, \hat{\mathbb{T}}$ are the empirical distributions of source domain and target domain, $n$ is the number target examples. \cite{zhang2021how} has proved that the original Mixup reduces $\gamma_{\mathcal{H}}(\mathbb{S}, \mathbb{T})$ by minimizing $\mathcal{R}(h|\mathbf{Z})$. Derived from the original Mixup, our mixup framework further minimizes $\gamma_{\mathcal{H}}(\hat{\mathbb{S}}, \hat{\mathbb{T}})$ by implicit domain alignment with the help of estimated $\mathbf{\hat{W}}$.

% \begin{equation}\mathcal{L}_{d i v}\left(f_{t} ; \mathcal{X}_{t}\right)=\sum_{k=1}^{K} \hat{p}_{k} \log \hat{p}_{k},\end{equation}
% where \begin{equation}\hat{p}_{k}=\mathbb{E}_{x^{t} \in \mathcal{X}_{u}^{t}}\left[S_{k} (f_{\theta_{c}}\left(f_{\theta_{e}}\left(x^{t}\right)\right) ) \right],
% \end{equation}
% Then the mixup samples $x_m$ and labels $y_m$ can be generated as follows:
% \begin{equation}
%    \begin{array}{ll}
%     x_{m}= \lambda x_{1} + (1-\lambda) x_{2};  x_{1} \in S, x_{2} \in S_{r}\\
%     y_{m}= \lambda y_{1} + (1-\lambda) y_{2};  y_{1} \in \hat{Y}, y_{2} \in \hat{Y}_{r}
%    \end{array}
% \end{equation}

\begin{algorithm}[tb]
    \caption{The Adaptation Stage of UIDM }
    \label{alg:algorithm1}
    \begin{flushleft}
    \textbf{Input}: The pre-trained encoder $f$ and transform matrix $\mathbf{\hat{W}}$,
    the labeled target data $\mathbf{X}_{l}^{t} $, $\mathbf{Y}_{l}^{t}$ and the unlabeled target data $\mathbf{X}_{u}^{t}$ \\
    \textbf{Output}: the updated encoder $f$.
    \end{flushleft}
    \begin{algorithmic}[1] %[1] enables line numbers
    \WHILE{not convergence}
    \STATE Compute uncertainty ${e_{i}}$ for each  unlabeled target example in $\mathbf{X}_{t}^{u}$ by  Equation \ref{cal_P}. 
    \STATE Compute soft labels for $\mathbf{X}_{t}^{u}$ and split them into low uncertainty set $\{\mathbf{I}, \mathbf{\hat{Y}}\}$ and high uncertainty set \{$\mathbf{ I}_{r},  \mathbf{\hat{Y}}_{r}\}$  using source-like selection.
    \FOR {$Step = 0 \to T$}
    \STATE Sample batch data $\mathbf{x} \in \mathbf{X}_{l}^{t} \cup   \mathbf{I}$, $\mathbf{y} \in \mathbf{Y}_{l}^{t} \cup   \mathbf{\hat{Y}}$ and $\mathbf{i} \in \mathbf{I}_{r}$, $\mathbf{y}_{r} \in \mathbf{\hat{Y}}_{r}$
%     \STATE Generate $\mathbf{x}_{a}, \mathbf{y}_{a}$ from shuffled set $\{ \mathbf{x} \cup \mathbf{i}, \mathbf{y} \cup \mathbf{y}_{r} \}$
    \STATE Generate interpolated examples $\{\mathbf{x},\mathbf{y}\}$ and $\{\mathbf{u}, \mathbf{y}_{u}\}$ using Equation \ref{LLL}, \ref{UUU}  and \ref{MMM}
    \STATE Update the encoder $f$ using $\{\mathbf{x},\mathbf{y}\}$ and $\{\mathbf{u}, \mathbf{y}_{u}\}$ with Equation \ref{LOSS}.
    \ENDFOR
    \ENDWHILE
    \end{algorithmic}
    \label{ALG}
    \end{algorithm}
% \subsection{why UUMIX work}
% \section{Data-dependent Generalization Bound for HTL}

\begin{table*}[bhtp]
% \begin{table*}[!hp]
   \centering
     \caption{The accuracies of semi-supervised domain adaptation on DomainNet dataset with ResNet34. 
   R: real, S: sketch, P: painting, C: clipart.
   We repeat three times per task and report the mean with a standard variance. The variance 0.0 denotes missing the value reported by the original paper.  
   }
   \begin{tabular}{c|ccccccc|c}
   \toprule
   Methods & R $\rightarrow$ S & R$\rightarrow$P & C$\rightarrow$S & R$\rightarrow$C 
   & P$\rightarrow$C & P$\rightarrow$R & S$\rightarrow$P & Mean \\ \hline
   
   \multicolumn{1}{l}{} & \multicolumn{8}{c}{1-Shot Adaptation} \\ \hline
   S+T   &$46.3_{\pm 0.0}$   & $60.6_{\pm 0.0}$ &$50.8_{\pm 0.0}$    &$55.6_{\pm 0.0}$ 
      &$56.8_{\pm 0.0}$      &$71.8_{\pm 0.0}$		 &$56.0_{\pm 0.0}$ 	   &$56.9_{\pm 0.0}$\\
   DANN  &$52.2_{\pm 0.0}$   &$61.4_{\pm 0.0}$  &$52.8_{\pm 0.0}$	   & $58.2_{\pm 0.0}$    &$56.3_{\pm 0.0}$ 
       	  &$70.3_{\pm 0.0}$		  &$57.4_{\pm 0.0}$ 	&$58.4_{\pm 0.0}$\\
   ADR      &$49.0_{\pm 0.0}$   &$61.3_{\pm 0.0}$  &$51.0_{\pm 0.0}$	  &  $57.1_{\pm 0.0}$   &$57.0_{\pm 0.0}$   
         &$72.0_{\pm 0.0}$		 &$56.0_{\pm 0.0}$ 	   &$57.6_{\pm 0.0}$\\ 
   CDAN  &$54.5_{\pm 0.0}$   &$64.9_{\pm 0.0}$  &$53.1_{\pm 0.0}$	   &$65.0_{\pm 0.0}$      &$63.7_{\pm 0.0}$    
       &$73.2_{\pm 0.0}$	  &$63.4_{\pm 0.0}$ 	&$62.5_{\pm 0.0}$\\
   SHOT & $60.4_{\pm 0.3}$ & $67.0_{\pm 0.3}$ & $61.0_{\pm 0.1}$ &$ 69.0_{\pm 0.1}$ & $69.4_{\pm 0.1}$ 
   & $\mathbf{79.4}_{\pm 0.1}$ & $62.4_{\pm 0.1}$ &$67.0_{\pm 0.1}$ \\ 
   ENT    &$52.1_{\pm 0.0}$   &$65.9_{\pm 0.0}$  &$54.6_{\pm 0.0}$    &$65.2_{\pm 0.0}$	     &$65.4_{\pm 0.0}$ 
          &$75.0_{\pm 0.0}$		 &$59.7_{\pm 0.0}$     &$62.6_{\pm 0.0}$\\
   MME    &$61.0_{\pm 0.0}$   &$67.7_{\pm 0.0}$  &$56.3_{\pm 0.0}$    &$70.0_{\pm 0.0}$	   &$69.0_{\pm 0.0}$  
      	 &$76.1_{\pm 0.0}$	   &$64.8_{\pm 0.0}$ 	&$66.4_{\pm 0.0}$\\
   APE & $63.0_{\pm 0.0}$ &$\mathbf{70.8}_{\pm 0.0}$ & $56.7_{\pm 0.0}$ & $70.4_{\pm 0.0}$ 
   & $72.9_{\pm 0.0}$ & $76.6_{\pm 0.0}$ & $64.5_{\pm 0.0}$ &$ 67.6_{\pm 0.0}$ \\
   BiAT     &$58.5_{\pm 0.0}$  &$68.0_{\pm 0.0}$ &$57.9_{\pm 0.0}$  &$73.0_{\pm 0.0}$ 	 &$71.6_{\pm 0.0}$	
   &$77.0_{\pm 0.0}$ &$63.9_{\pm 0.0}$  &$67.1_{\pm 0.0}$\\ \hline

   % S+T & 50.1 & 62.2 & 55.0 & 60.0 & 59.4 & 73.9 & 59.5 & 60.0 \\ 
   % DANN & 54.9 & 62.8 & 55.4 & 59.8 & 59.6 & 72.2 & 59.9 & 60.7 \\ 
   % ADR & 51.1 & 61.9 & 54.4 & 60.7 & 60.7 & 74.2 & 59.9 & 60.4 \\ 
   % CDAN & 59.0 & 67.3 & 57.8 & 69.0 & 68.4 & 78.5 & 65.3 & 66.5 \\ 
   % SHOT & 59.3 & 66.5 & 61.2 & 68.7 & 69.3 & 80.0 & 63.4 & 66.9 \\ 
   % ENT & 61.1 & 69.2 & 60.0 & 71.0 & 71.1 & \textbf{78.6} & 62.1 & 67.6 \\ 
   % MME & 61.9 & 69.7 & 61.8 & 72.2 & 71.7 & 78.5 & 66.8 & 68.9 \\ 
   % APE & 62.1 & 68.8 & 61.5 & \textbf{74.9} & 74.6 & \textbf{78.6} & \textbf{67.5} & 69.7 \\ 
   % Meta & 62.1 & 68.8 & 61.5 & \textbf{74.9} & 74.6 & \textbf{78.6} & \textbf{67.5} & 69.7 \\ 
   % BiAT & 62.1 & 68.8 & 61.5 & \textbf{74.9} & 74.6 & \textbf{78.6} & \textbf{67.5} & 69.7 \\ \hline
   
   UIDM (Ours) 
   & ${\mathbf{63.5}}_{\pm 0.3}$
   & ${{69.2}}_{\pm 0.3}$
   & ${\mathbf{64.5}}_{\pm 1.0}$
   & ${\mathbf{74.1}}_{\pm 1.5}$
   & ${\mathbf{73.3}}_{\pm 1.1}$
   & ${{74.5}}_{\pm 0.8}$
   & ${\mathbf{67.0}}_{\pm 0.5}$
   & ${\mathbf{69.5}}_{\pm 0.4}$
   \\ \midrule

   \multicolumn{1}{l}{} & \multicolumn{8}{c}{3-Shot Adaptation} \\ \hline
   S+T & $50.1_{\pm 0.0}$ &  $62.2_{\pm 0.0}$ &  $55.0_{\pm 0.0}$ & $ 60.0_{\pm 0.0}$ 
   &  $59.4_{\pm 0.0}$ & $ 73.9_{\pm 0.0}$ & $ 59.5_{\pm 0.0}$ & $ 60.0_{\pm 0.0}$ \\
   DANN & $ 54.9_{\pm 0.0}$ & $ 62.8_{\pm 0.0}$ & $ 55.4_{\pm 0.0}$ & $ 59.8_{\pm 0.0}$ 
   & $ 59.6_{\pm 0.0}$ & $ 72.2_{\pm 0.0}$ & $ 59.9_{\pm 0.0}$ & $60.7_{\pm 0.0}$ \\ 
   ADR &  $51.1_{\pm 0.0}$ &  $61.9_{\pm 0.0}$ & $ 54.4_{\pm 0.0}$ &  $60.7_{\pm 0.0}$ 
   & $ 60.7_{\pm 0.0}$ & $ 74.2_{\pm 0.0}$ &  $59.9_{\pm 0.0}$ & $ 60.4_{\pm 0.0}$ \\ 
   CDAN &  $59.0_{\pm 0.0}$ &  $67.3_{\pm 0.0}$ &  $57.8_{\pm 0.0}$ &  $69.0_{\pm 0.0}$ 
   &  $68.4_{\pm 0.0}$ &  $78.5_{\pm 0.0}$ &  $65.3_{\pm 0.0}$ & $ 66.5_{\pm 0.0}$ \\ 
   SHOT & $ 59.3_{\pm 0.5}$ &  $66.5_{\pm 2.6}$ & $ 61.2_{\pm 2.5}$ & $ 69.2_{\pm 0.4}$ 
   &  $69.6_{\pm 0.2}$ & $ \mathbf{80.0}_{\pm 0.0}$ &  $63.4_{\pm 6.3}$ &  $67.2_{\pm 0.8}$ \\ 
   ENT &  $61.1_{\pm 0.0}$ & $ 69.2_{\pm 0.0}$ & $ 60.0_{\pm 0.0}$ & $71.0_{\pm 0.0}$ 
   &  $71.1_{\pm 0.0}$ & $ {78.6}_{\pm 0.0}$ &  $62.1_{\pm 0.0}$ &  $67.6_{\pm 0.0}$ \\ 
   MME & $ 61.9_{\pm 0.0}$ &  $69.7_{\pm 0.0}$ & $ 61.8_{\pm 0.0}$ &  $72.2_{\pm 0.0}$ 
   & $ 71.7_{\pm 0.0}$ &  $78.5_{\pm 0.0}$ & $ 66.8_{\pm 0.0}$ & $ 68.9_{\pm 0.0}$ \\ 
   BiAT & $ 62.1_{\pm 0.0}$ & $ 68.8_{\pm 0.0}$ &  $61.5_{\pm 0.0}$ 
   &  ${74.9}_{\pm 0.0}$ &  $74.6_{\pm 0.0}$ &  ${78.6}_{\pm 0.0}$ 
   &  ${67.5}_{\pm 0.0}$ & $ 69.7_{\pm 0.0}$ \\ 
   Meta-MME & $63.8_{\pm 0.0}$ & $ 70.3_{\pm 0.0}$ & $62.8_{\pm 0.0}$ & $73.5_{\pm 0.0}$ 
   &$72.8_{\pm 0.0}$ & $79.2_{\pm 0.0}$ & $68.0_{\pm 0.0}$ & $70.1_{\pm 0.0}$ \\
   APE & $\mathbf{66.2}_{\pm 2.1}$ & $\mathbf{71.2}_{\pm 0.5}$ & $63.9_{ \pm 1.2}$ &  
   ${72.9}_{\pm 2.2} $ & $72.7_{ \pm  0.1}$ 
   & ${78.0}_{\pm 1.5}$ & ${67.9}_ {\pm 1.1}$ & ${70.4}_{\pm 0.3} $ \\\hline
   UIDM (Ours) 
   & ${{65.5}}_{\pm 0.1}$
   & ${{70.3}}_{\pm 0.6}$
   & ${\mathbf{65.2}}_{\pm 0.1}$
   & ${\mathbf{75.5}}_{\pm 0.4}$
   & ${\mathbf{75.5}}_{\pm 0.3}$
   & ${{78.8}}_{\pm 0.1}$
   & ${\mathbf{68.2}}_{\pm 0.1}$
   & ${\mathbf{71.2}}_{\pm 0.1}$
   \\ \bottomrule
   \end{tabular}
 
\label{tab:domainNet}
   \end{table*}
   
\section{Experiments}
\subsection{Datasets}
We adopt three public datasets including \textbf{DomainNet} \cite{peng2018moment},  \textbf{Office-Home} \cite{venkateswara2017Deep}  and  \textbf{Office-31} \cite{office}.

 \textbf{DomainNet } is a large dataset containing six domains including Clipart, Infograph, Painting, 
 Real, Quickdraw \cite{peng2018moment}.  Following MME \cite{MME}, 
  we also pick 4 domains (Real, Clipart, Painting, Sketch), and 126 classes for each domain. 
  
   \textbf{Office-Home}  consists of images from 4 domains: Art, Clipart, Product and Real-World images. For each domain, the dataset contains images of 65 object categories built typically from office and home environments. 
   
 \textbf{Office-31}  dataset is a small size dataset containing 3 domains (Amazon, Webcam and Dslr) and 31 categories in each domain.

%   The Amazon domain contains images from the Amazon website, and the Webcam and Dslr domain contain the office environment images taken with varying lighting and pose changes using a webcam and dslr camera, respectively. 
%  ADR adopts the dropout scheme to modify the decision boundary for feature alignment. CDAN adversarially aligns the feature by fooling the conditional domain discriminator.

\subsection{Baselines}
The baselines including three categories: \textbf{None Adaptation} (S+T \cite{MME}), \textbf{Semi-Supervised Learning} (ENT \cite{EM,MME}),
 \textbf{Unsupervised Domain Adaptation} (DANN \cite{ganin2015unsupervised}, ADR \cite{saito2018adversarial}, 
 CDAN \cite{long2018conditional}, SHOT \cite{SHOT}), and \textbf{Semi-Supervised Domain Adaptation} (MME \cite{MME}, BiAT \cite{BiAT}, Meta-MME \cite{li2020online}, APE \cite{APE}).
 For the unsupervised domain adaptation methods, the labeled target data were put with the source data during the adaptation process.  
% PAE \cite{kim2020attract}, OML \cite{li2020online}
S+T trains a model with labeled source and labeled target data without using unlabeled 
data from target domain.  ENT is a direct semi-supervised method that minimizes the entropy of the 
unlabeled target data. 
MME proposed minimax entropy loss for the SSDA problem. Based on MME, Meta-MME achieves better generalization with the help of Meta Learning \cite{pmlr-v70-finn17a, ASMAML}.
BiAT uses bidirectional adversarial 
training for generating samples between the source and target domain. APE uses data perturbation and 
exploration to reduce the intra-domain discrepancy. 
We reproduce the APE method with model selection according to the original code, in which the model selection process is ignored. Besides, SHOT is an unsupervised domain adaptation methods via Source Hypothesis Transfer. We also reproduce SHOT method including the semi-supervised and the unsupervised version from their original codes. 

\subsection{Implementation Details}
\textbf{Data preparation. } At the pre-training stage, we split the source data  into the training data and the validation data, because of no access to the target domain. Specifically, on Office-31 and Office-Home, we split the training data and the validation data as 0.9 : 0.1;  on DomainNet, we split the training data and the validation data as 0.98 : 0.02. We use the standard data augmentation 
methods consisting of random horizontal flip, random crop,  and data normalization. 
For a fair comparison with baselines, we use the same partition for the target domain 
data as \cite{MME} corresponding to the labeled training data, the unlabeled training data, 
and the labeled validation data.  Except for the final performance evaluation stage on the unlabeled data, 
we also use the same data augmentation methods when performing adaptation.

\textbf{Model Architecture. }
We select three encoder backbones including AlexNet \cite{krizhevsky2012imagenet}, VGGNet \cite{simonyan2014very} and the  ResNet-34 \cite{he2016deep}. For the AlexNet and the VGGNet, we add a bottleneck layer after the last layer of the encoder.  For the ResNet-34, we drop the last layer of the model and add a  bottleneck layer like the former backbones.   Finally,  we use a classifier with one normalized fully connected layer. 

\textbf{Training Setting. }
% On the training stage, we set the learning rate as 0.001 for the encoder except the last layer with 0.005.  We also set the learning rate of the classifier as 0.005. 
We set the learning rate as 0.001 for all encoders and 0.01 for the rest layers.   According to validation set, the selected numbers of examples per class for  Office-31, Office-home, DomainNet are $\{3,5,20\}$.  Following SHOT, we freeze the classifier at the adaptation stage and apply entropy constraints for the smaller dataset Office-31 and Office-Home. Furthermore, for all datasets, the mixup coefficients $\lambda$ is sampled from a Beta Distribution $Beta(2, 0.5)$,  the weight $\alpha$ of $ \mathcal{L}_{mse}$  is set to $200$.  The repeat times $n_{r}$ is empirically set to 5.
% For the weights of different  regularization items, we find that 1, 0.1, 1,1 is good enough for most of adaptation experiments. Note that compared to SHOT, we also apply entropy constraints on the two 
% smaller datasets.
For all experiments, we randomly select three different seeds to repeat the experiments and  report the mean results with a standard variance.  Besides, our experiment was implemented with Pytorch \footnote{https://pytorch.org/} and running on one  RTX 3090 GPU. 
% DANN performed domain distributions alignment through adversarial lea
   \begin{table}[]
   \centering
    \caption{The accuracies of semi-supervised domain adaptation on Office-31 dataset with AlexNet and Office-Home
      dataset VGGNet. The missing values are denoted by '-'. On Office-31, we focus on  Dslr to Amazon and Webcam to Amazon. 
      On Office-Home, 12 adaptation tasks are preformed across 4 domains.}
      \begin{tabular}{c|cc|cc}
      \toprule
      \multicolumn{1}{c}{Methods} & \multicolumn{2}{c}{Office-31} & \multicolumn{2}{c}{Office-Home} \\ \hline
       & 1-shot & 3-shot & 1-shot & 3-shot \\ \hline
      S+T & $50.2_{\pm 0.0}$ & $61.8_{\pm 0.0}$ & $57.4_{\pm 0.0}$ & $62.9_{\pm 0.0}$ \\
      DANN & $55.8_{\pm 0.0}$ & $64.8_{\pm 0.0}$ & $60.0_{\pm 0.0}$ &$ 63.9_{\pm 0.0}$ \\
      ADR & $50.6_{\pm 0.0}$ & $61.3_{\pm 0.0}$ &$ 57.4_{\pm 0.0}$ & $63.0_{\pm 0.0}$ \\
      CDAN &$ 49.4_{\pm 0.0}$ & $60.8_{\pm 0.0}$ & $55.8_{\pm 0.0}$ &$ 61.8_{\pm 0.0}$ \\
      SHOT &$ 51.7_{\pm 2.7}$ & $63.3_{\pm 0.0}$ &$ 59.9_{\pm 0.1}$ &$ 66.2_{\pm 0.1}$ \\
      ENT &$ 48.1_{\pm 0.0}$ & $65.5_{\pm 0.0}$ &$ 51.6_{\pm 0.0}$ & $64.8_{\pm 0.0}$ \\
      MME &$ 56.5_{\pm 0.0}$ & $67.5_{\pm 0.0}$ & $62.7_{\pm 0.0}$ & $67.6_{\pm 0.0}$ \\
      BiAT & $56.2_{\pm 0.0}$ & $\mathbf{68.3}_{\pm 0.0}$ & - & - \\ \hline 
      UIDM & $\mathbf{58.3}_{\pm 0.4}$ & $68.2_{\pm 0.3}$ & $\mathbf{64.3}_{\pm 0.1}$ 
      & $\mathbf{69.1}_{\pm 0.2}$\\ 
      \bottomrule
      \end{tabular}
     
      \label{tab:office}
      \end{table}

%  \begin{figure*}[!hbtp]
 \begin{figure*}[!t]
\centering
\subfigure[\scriptsize{Domain Level without Adaptation}]{
\label{fig:com1}
\centering
\includegraphics[width=0.215\textwidth]{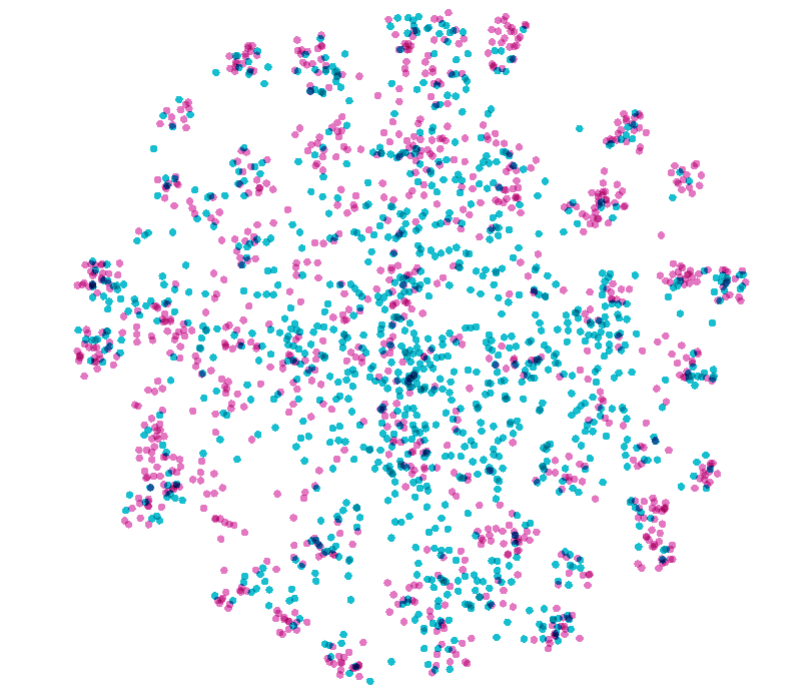}
}
% \hspace{.2in}
\subfigure[\scriptsize{Domain Level with Adaptation}]{
\label{fig:com2}
\centering
\includegraphics[width=0.225\textwidth]{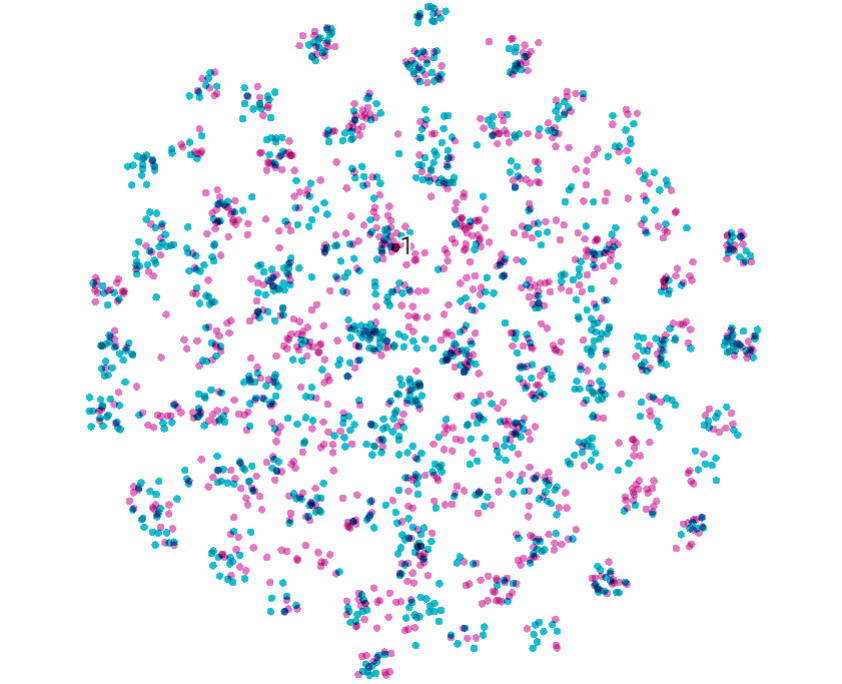}
}
% \hspace{.2in}
\subfigure[\scriptsize{Class Level without Adaptation}]{
\label{fig:com3}
\centering
\includegraphics[width=0.23\textwidth]{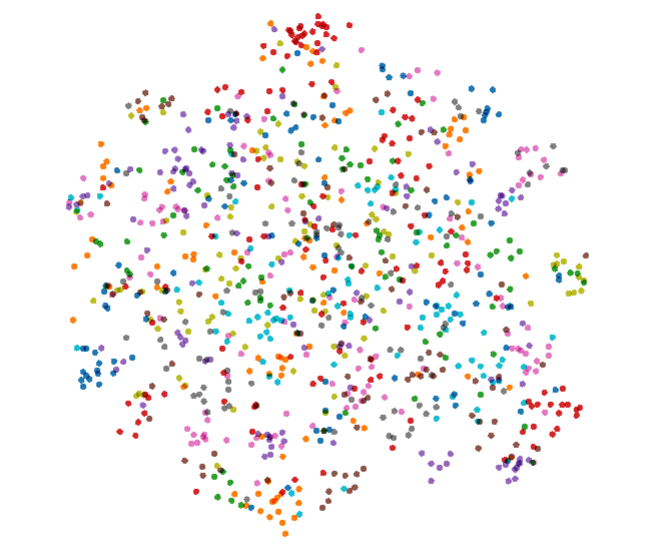}
}
% \hfill
\subfigure[\scriptsize{Class Level with Adaptation}]{
\label{fig:example4}
\centering
\includegraphics[width=0.21\textwidth]{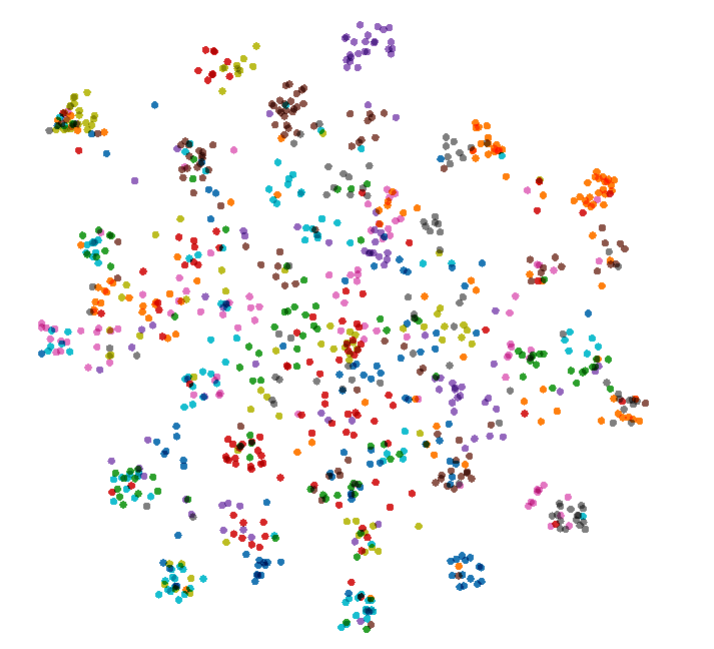}
}
\caption{T-SNE visualization for the encoder's outputs of 
the source domain Real and the target domain Sketch on DomainNet.
For clarity, we randomly select 1000 examples from each domain.
 (a) The visualization of the source domain and target domain using the model trained on the source domain. The \textcolor{magenta}{magenta}  dots denote the data of the source domain Real, and the \textcolor{cyan}{cyan} dots denote the data of the target domain Sketch. (b) After adaptation, the target domain's representation becomes more compact and is clustered to the source domain.
 (c) The class level visualization of the target domain Sketch before adaptation. Each class is denoted by one color.
 (d) After adaptation, the representation of each class is more compact.}

\label{fig:train_process}
\end{figure*} 
\subsection{Results}
We focus on the evaluation of 1-shot/3-shot domain adaptation on the large scale dataset DomainNet (see Table \ref{tab:domainNet} and Figure \ref{fig:train_process}). 
Our method outperforms the others on most semi-supervised domain adaptation tasks. Compared to the BiAT method that generates
adversarial examples between two domains, our method performs target adaptation without accessing the source domain but 
gets stronger performance by generating interpolated examples only in the target domain. Similarly, APE uses data perturbation and 
exploration to reduce the intra-domain discrepancy.  Our mixup strategy can effectively explore the data structure compared to data perturbation and exploration strategy.

Compared to SHOT (see Table \ref{tab:domainNet}), which is an unsupervised domain adaptation method proposed with privacy preserving consideration.
Our method outperforms SHOT on all the tasks except the simple task Painting to Real. The reason is that the diversity-promoting objective in SHOT is suboptimal when dealing with large scale dataset. Because this objective has multiple solutions which bring more uncertainty for the optimization as the class increasing.
Recent work for multi-source domain adaptation \cite{feng2021kd3a}
also verifies that SHOT is suboptimal on large scale dataset like DomainNet. We also conduct unsupervised domain adaptation
using SHOT and our unsupervised version (See Table \ref{tab:uda}). Under the unsupervised setting, no labeled target examples are available. Instead, we use the 
selected target examples with low uncertainty to replace the labeled target examples.  
\begin{table}[!tph]
    \centering
     \caption{
      Performances of unsupervised domain adaptation on DomainNet with ResNet34. 
      No labeled target examples are used in the adaptation of UIDM and SHOT. Instead, we use the 
      selected traget examples with low uncertainty to replace the labeled target examples.    
   }
   \begin{tabular}{l|ccc}
   \toprule
   \multicolumn{1}{c}{Methods}  & C$\rightarrow$S & R $\rightarrow$ S  & S$\rightarrow$P
      \\ \hline
   Source-Only  & $49.2_{\pm 0.5}$ & $54.6_{\pm 0.3}$   & $51.3_{\pm 0.4}$\\ \hline
   
   SHOT  & $60.3_{\pm 0.1}$ &$ 59.3_{\pm 0.1}$  & $64.2_{\pm 0.1}$ \\ \hline
   UIDM   (Ours) & $\mathbf{63.9}_{\pm 0.5}$ & $\mathbf{63.5}_{\pm 0.1}$  &$\mathbf{ 66.0}_{\pm 0.2}$\\ 
   
 \midrule
   \multicolumn{1}{c}{Methods} & R$\rightarrow$C & R$\rightarrow$P & P $\rightarrow$ C  
      \\ \hline
   Source-Only & $58.3_{\pm 0.2}$ & $50.2_{\pm 0.3}$ & $57.6_{\pm 0.3}$   \\ \hline
   
   SHOT & $66.9_{\pm 0.1}$  &$ 63.6_{\pm 0.1}$  & $68.1_{\pm 0.3}$

   \\ \hline
   UIDM (Ours) &$ \mathbf{72.9}_{\pm 0.2}$ & $\mathbf{69.5}_{\pm 0.3}$ & $\mathbf{71.7}_{\pm 0.1}$  \\
   
   \bottomrule
   \end{tabular}
  
   \label{tab:uda}
   \end{table} 
   
For Office-31 and Office-Home, recent baselines such as BiAT and Meta-MME lack uniform evaluation on the two datasets. The average performance for 1-shot/3-shot adaptation tasks 
can be seen in Table \ref{tab:office}. 

\subsection{Ablation Study}
Table \ref{tab:ablation} demonstrates different performances by removing multiple operations in turn. 
When removing the source-like selection, our method equals to original Mixmatch \cite{MixMatch}, which overfits on 
1-shot domain adaptation tasks. When hybrid-mixup is removed, the performance also suffers large degradation, because 
only performing mixup on the unlabeled data can not performing efficient domain alignment. Besides, self-mixup has smaller contribution to the overall performance.

\begin{table}[!tph]
\centering
      \caption{
       Ablation results of 1-shot adaptation on DomainNet with ResNet34. Source-Only denotes not using adaptation and evaluating with pre-trained models. Selection hybrid-M and Self-M denote source-like selection. 
       hybrid mixup operation and self-mixup operation. 
       }
      \begin{tabular}{l|ccc}
      \toprule
      \multicolumn{1}{c}{Methods} & S$\rightarrow$P & C$\rightarrow$S & R $\rightarrow$ S  
         \\ \hline
      Source-Only & $51.3_{\pm 0.4}$ & $49.2_{\pm 0.5}$ & $54.6_{\pm 0.3}$   \\ \hline
      UIDM w/o Selection & $52.4_{\pm 0.5}$ & $45.6_{\pm 1.4}$ & $51.5_{\pm 0.8}$  \\
      UIDM w/o hybrid-M & $58.8_{\pm 0.1}$ & $53.5_{\pm 0.1}$ &$ 53.2_{\pm 0.1}$\\
      UIDM w/o Self-M & $62.3_{\pm 0.3}$ & $58.8_{\pm 0.2}$ &$ 60.5_{\pm 0.1}$  

      \\ \hline
      UIDM full &$ \mathbf{67.0}_{\pm 0.2}$ & $\mathbf{64.5}_{\pm 0.6}$ & $\mathbf{63.5}_{\pm 0.1}$  \\
      \bottomrule
      \end{tabular}

       \label{tab:ablation}
   \end{table}

   \begin{figure}[t]
      \begin{center}
      \centering
      \includegraphics[width=0.99\linewidth]{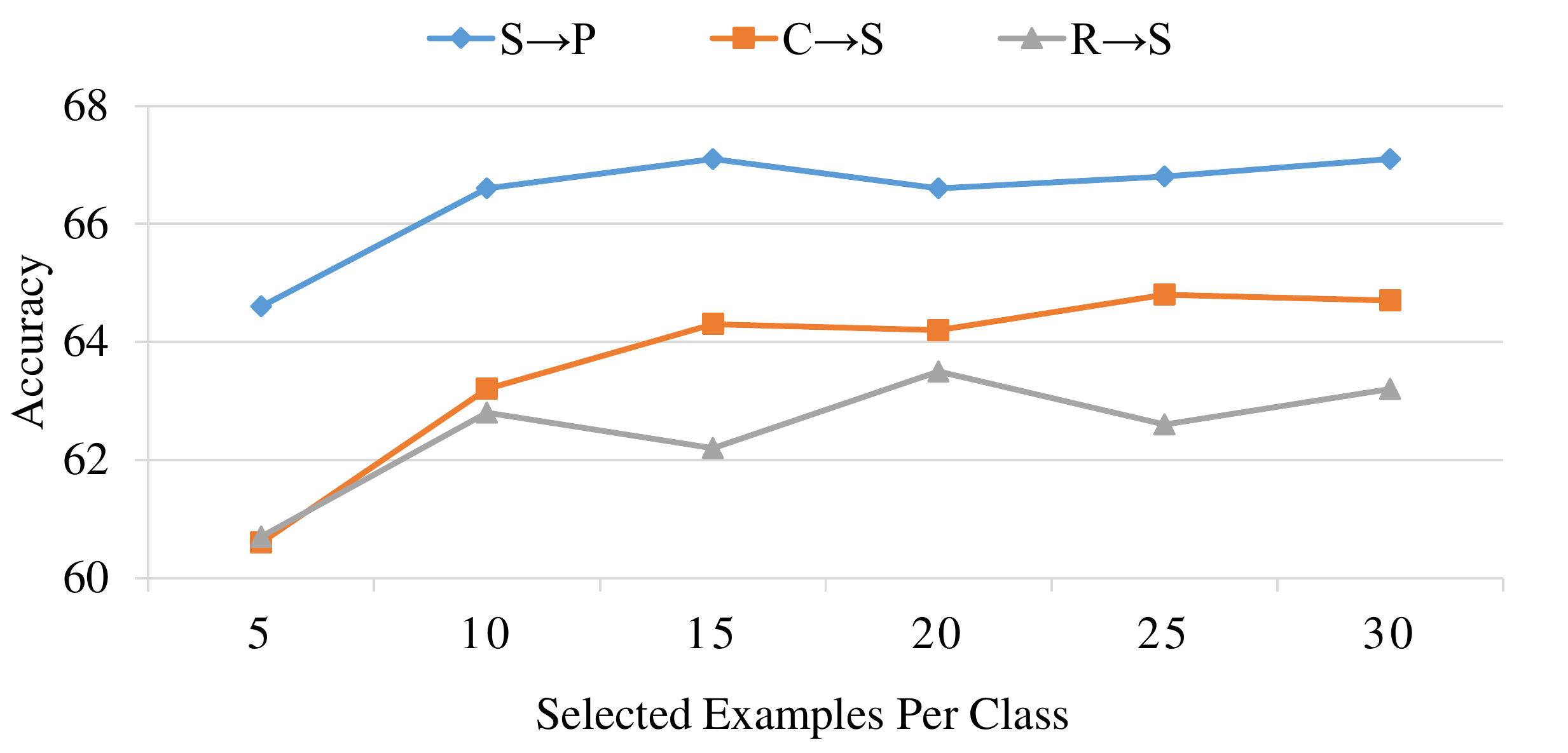}
      \end{center}
         \caption{The performances when selecting different number of examples per class on DomainNet with ResNet34.}
      \label{fig:SNPC}
      \end{figure}

   \begin{figure}[t]
   \centering
      \begin{center}
      \includegraphics[width=0.99\linewidth]{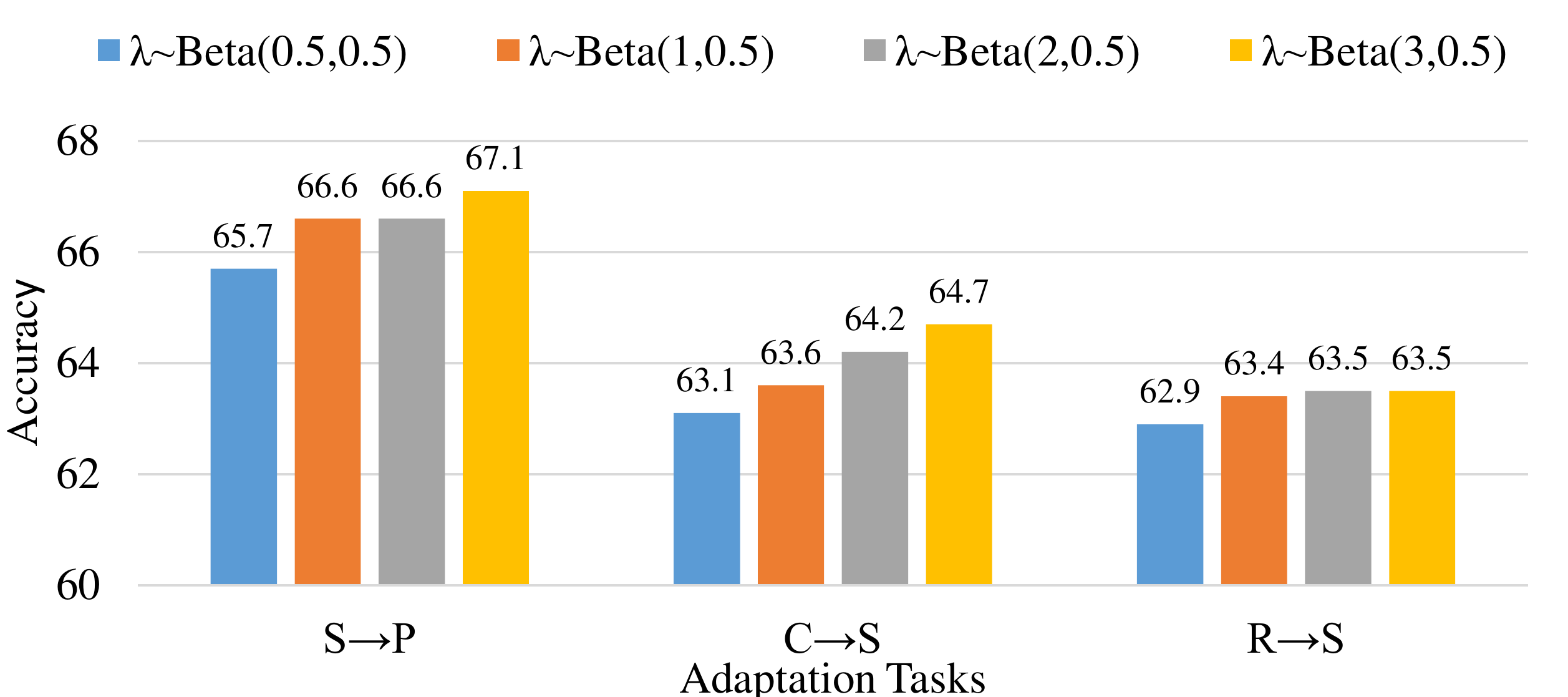}
      \end{center}
         \caption{The performances when selecting different mixup coefficient $\lambda$ on DomainNet with ResNet34.}
      \label{fig:lambda}
      \end{figure}
      
\subsection{Sensitivity Analysis}
   The selected number of example per class (SNPC) and the mixup coefficient $\lambda$  are critical hyper-parameters for our framework. 

   To demonstrate the robustness of our method on different SNPC, we conduct 
   three adaptation tasks on large-scale dataset DomainNet, given multiple SNPC varying 5 to 30. 
   Figure \ref{fig:SNPC} demonstrates the adaptation accuracies corresponding to each setting. 
   Compared to SNPC=0,  SNPC=5 improves the performance with a large margin.  When given larger SNPC, the performance is becoming flattened. Even so, we hope more elegant way to decide the selection of SNPC and we leave it as future work.

   To demonstrate the robustness of our method on different mixup up coefficient $\lambda$, we sample 
   $\lambda$ from a Beta distribution keep its mean more than $0.5$. From Figure \ref{fig:lambda}, we can 
   see when the mean of $\lambda$ is close to $0.5$, the performance degradation appears. Because the 
   network is incapable to decide which part is important in mixup.  When $\lambda > 0.5$, the performance is improved with acceptable standard variance.

\section{Conclusion and Future Work}
In this paper, we propose Uncertainty-guided Intra-Domain Mixup(UIDM) method to perform semi-supervised 
target domain adaptation without accessing the  source data. With the help of uncertainty estimation,
we split the target examples into three groups: the unlabeled target examples with high uncertainty, 
the unlabeled target examples with low uncertainty and the labeled target examples.
Then Self-Mixup and Hybrid-Mixup are used to perform intra-group and inter-group data interpolation, respectively. Compared to standard mixup methods like Mixmatch, our framework still works well when only given only 
one or three labeled examples in virtue of the mechanism of source-like selection. In the future, 
we hope to select optimal source-like examples for each class to improve the semi-supervised and even unsupervised domain adaptation under source-free setting.

{\small
\bibliographystyle{ieee_fullname}
\bibliography{egpaper_final}

\begin{thebibliography}{10}\itemsep=-1pt

\bibitem{RRR}
Peter~L. Bartlett and Shahar Mendelson.
\newblock Rademacher and gaussian complexities: Risk bounds and structural
  results.
\newblock {\em J. Mach. Learn. Res.}, 3(null):463–482, Mar. 2003.

\bibitem{MixMatch}
David Berthelot, Nicholas Carlini, Ian Goodfellow, Nicolas Papernot, Avital
  Oliver, and Colin~A Raffel.
\newblock Mixmatch: A holistic approach to semi-supervised learning.
\newblock In {\em Advances in Neural Information Processing Systems 32}, pages
  5049--5059. Curran Associates, Inc., 2019.

\bibitem{blundell15}
Charles Blundell, Julien Cornebise, Koray Kavukcuoglu, and Daan Wierstra.
\newblock Weight uncertainty in neural network.
\newblock In Francis Bach and David Blei, editors, {\em Proceedings of the 32nd
  International Conference on Machine Learning}, volume~37 of {\em Proceedings
  of Machine Learning Research}, pages 1613--1622, Lille, France, 07--09 Jul
  2015. PMLR.

\bibitem{kddsourcefree}
Boris Chidlovskii, Stephane Clinchant, and Gabriela Csurka.
\newblock Domain adaptation in the absence of source domain data.
\newblock In {\em Proceedings of the 22nd ACM SIGKDD International Conference
  on Knowledge Discovery and Data Mining}, KDD ’16, page 451–460, New York,
  NY, USA, 2016. Association for Computing Machinery.

\bibitem{feng2021kd3a}
Hao-Zhe Feng, Zhaoyang You, Minghao Chen, Tianye Zhang, Minfeng Zhu, Fei Wu,
  Chao Wu, and Wei Chen.
\newblock Kd3a: Unsupervised multi-source decentralized domain adaptation via
  knowledge distillation, 2021.

\bibitem{pmlr-v70-finn17a}
Chelsea Finn, Pieter Abbeel, and Sergey Levine.
\newblock Model-agnostic meta-learning for fast adaptation of deep networks.
\newblock In Doina Precup and Yee~Whye Teh, editors, {\em Proceedings of the
  34th International Conference on Machine Learning}, volume~70 of {\em
  Proceedings of Machine Learning Research}, pages 1126--1135, International
  Convention Centre, Sydney, Australia, 06--11 Aug 2017. PMLR.

\bibitem{pmlr-v48-gal16}
Yarin Gal and Zoubin Ghahramani.
\newblock Dropout as a bayesian approximation: Representing model uncertainty
  in deep learning.
\newblock In Maria~Florina Balcan and Kilian~Q. Weinberger, editors, {\em
  Proceedings of The 33rd International Conference on Machine Learning},
  volume~48 of {\em Proceedings of Machine Learning Research}, pages
  1050--1059, New York, New York, USA, 20--22 Jun 2016. PMLR.

\bibitem{ganin2015unsupervised}
Yaroslav Ganin and Victor Lempitsky.
\newblock Unsupervised domain adaptation by backpropagation.
\newblock In {\em International conference on machine learning}, pages
  1180--1189, 2015.

\bibitem{GAN}
Ian Goodfellow, Jean Pouget-Abadie, Mehdi Mirza, Bing Xu, David Warde-Farley,
  Sherjil Ozair, Aaron Courville, and Yoshua Bengio.
\newblock Generative adversarial nets.
\newblock In Z. Ghahramani, M. Welling, C. Cortes, N. Lawrence, and K.~Q.
  Weinberger, editors, {\em Advances in Neural Information Processing Systems},
  volume~27. Curran Associates, Inc., 2014.

\bibitem{goodfellow2014generative}
Ian~J. Goodfellow, Jean Pouget-Abadie, Mehdi Mirza, Bing Xu, David
  Warde-Farley, Sherjil Ozair, Aaron Courville, and Yoshua Bengio.
\newblock Generative adversarial networks, 2014.

\bibitem{EM}
Yves Grandvalet and Yoshua Bengio.
\newblock Semi-supervised learning by entropy minimization.
\newblock In {\em Advances in neural information processing systems}, pages
  529--536, 2005.

\bibitem{he2016deep}
Kaiming He, Xiangyu Zhang, Shaoqing Ren, and Jian Sun.
\newblock Deep residual learning for image recognition.
\newblock In {\em Proceedings of the IEEE conference on computer vision and
  pattern recognition}, pages 770--778, 2016.

\bibitem{BiAT}
Pin Jiang, Aming Wu, Yahong Han, Yunfeng Shao, Meiyu Qi, and Bingshuai Li.
\newblock Bidirectional adversarial training for semi-supervised domain
  adaptation.
\newblock In Christian Bessiere, editor, {\em Proceedings of the Twenty-Ninth
  International Joint Conference on Artificial Intelligence, {IJCAI-20}}, pages
  934--940. International Joint Conferences on Artificial Intelligence
  Organization, 7 2020.
\newblock Main track.

\bibitem{wtat}
Alex Kendall and Yarin Gal.
\newblock What uncertainties do we need in bayesian deep learning for computer
  vision?
\newblock In {\em Proceedings of the 31st International Conference on Neural
  Information Processing Systems}, NIPS'17, page 5580–5590, Red Hook, NY,
  USA, 2017. Curran Associates Inc.

\bibitem{APE}
Taekyung Kim and Changick Kim.
\newblock Attract, perturb, and explore: Learning a feature alignment network
  for semi-supervised domain adaptation.
\newblock {\em ECCV}, 2020.

\bibitem{krizhevsky2012imagenet}
Alex Krizhevsky, Ilya Sutskever, and Geoffrey~E Hinton.
\newblock Imagenet classification with deep convolutional neural networks.
\newblock In {\em Advances in neural information processing systems}, pages
  1097--1105, 2012.

\bibitem{NIPS20104009}
Abhishek Kumar, Avishek Saha, and Hal Daume.
\newblock Co-regularization based semi-supervised domain adaptation.
\newblock In J.~D. Lafferty, C.~K.~I. Williams, J. Shawe-Taylor, R.~S. Zemel,
  and A. Culotta, editors, {\em Advances in Neural Information Processing
  Systems 23}, pages 478--486. Curran Associates, Inc., 2010.

\bibitem{li2020online}
Da Li and Timothy Hospedales.
\newblock Online meta-learning for multi-source and semi-supervised domain
  adaptation.
\newblock {\em ECCV}, 2020.

\bibitem{Li_2020_CVPR}
Rui Li, Qianfen Jiao, Wenming Cao, Hau-San Wong, and Si Wu.
\newblock Model adaptation: Unsupervised domain adaptation without source data.
\newblock In {\em Proceedings of the IEEE/CVF Conference on Computer Vision and
  Pattern Recognition (CVPR)}, June 2020.

\bibitem{SHOT}
Jian Liang, Dapeng Hu, and Jiashi Feng.
\newblock Do we really need to access the source data? source hypothesis
  transfer for unsupervised domain adaptation.
\newblock {\em ICML}, 2020.

\bibitem{long2018conditional}
Mingsheng Long, Zhangjie Cao, Jianmin Wang, and Michael~I Jordan.
\newblock Conditional adversarial domain adaptation.
\newblock In {\em Advances in Neural Information Processing Systems}, pages
  1640--1650, 2018.

\bibitem{ASMAML}
Ning Ma, Jiajun Bu, Jieyu Yang, Zhen Zhang, Chengwei Yao, Zhi Yu, Sheng Zhou,
  and Xifeng Yan.
\newblock Adaptive-step graph meta-learner for few-shot graph classification.
\newblock In {\em Proceedings of the 29th ACM International Conference on
  Information amp; Knowledge Management}, CIKM '20, page 1055–1064, New York,
  NY, USA, 2020. Association for Computing Machinery.

\bibitem{mishra2021surprisingly}
Samarth Mishra, Kate Saenko, and Venkatesh Saligrama.
\newblock Surprisingly simple semi-supervised domain adaptation with
  pretraining and consistency, 2021.

\bibitem{VAT}
Takeru Miyato, Shin-ichi Maeda, Masanori Koyama, and Shin Ishii.
\newblock Virtual adversarial training: a regularization method for supervised
  and semi-supervised learning.
\newblock {\em IEEE transactions on pattern analysis and machine intelligence},
  41(8):1979--1993, 2018.

\bibitem{IPM}
Alfred Müller.
\newblock Integral probability metrics and their generating classes of
  functions.
\newblock {\em Advances in Applied Probability}, 29(2):429--443, 1997.

\bibitem{peng2018moment}
Xingchao Peng, Qinxun Bai, Xide Xia, Zijun Huang, Kate Saenko, and Bo Wang.
\newblock Moment matching for multi-source domain adaptation.
\newblock {\em ICCV}, 2019.

\bibitem{Peng2020Federated}
Xingchao Peng, Zijun Huang, Yizhe Zhu, and Kate Saenko.
\newblock Federated adversarial domain adaptation.
\newblock In {\em International Conference on Learning Representations}, 2020.

\bibitem{rizve2021in}
Mamshad~Nayeem Rizve, Kevin Duarte, Yogesh~S Rawat, and Mubarak Shah.
\newblock In defense of pseudo-labeling: An uncertainty-aware pseudo-label
  selection framework for semi-supervised learning.
\newblock In {\em International Conference on Learning Representations}, 2021.

\bibitem{office}
Kate Saenko, Brian Kulis, Mario Fritz, and Trevor Darrell.
\newblock Adapting visual category models to new domains.
\newblock In {\em Proceedings of the 11th European Conference on Computer
  Vision: Part IV}, ECCV'10, page 213–226, Berlin, Heidelberg, 2010.
  Springer-Verlag.

\bibitem{MME}
Kuniaki Saito, Donghyun Kim, Stan Sclaroff, Trevor Darrell, and Kate Saenko.
\newblock Semi-supervised domain adaptation via minimax entropy.
\newblock {\em ICCV}, 2019.

\bibitem{saito2018adversarial}
Kuniaki Saito, Yoshitaka Ushiku, Tatsuya Harada, and Kate Saenko.
\newblock Adversarial dropout regularization.
\newblock In {\em International Conference on Learning Representations}, 2018.

\bibitem{simonyan2014very}
Karen Simonyan and Andrew Zisserman.
\newblock Very deep convolutional networks for large-scale image recognition.
\newblock {\em arXiv preprint arXiv:1409.1556}, 2014.

\bibitem{tarvainen2017mean}
Antti Tarvainen and Harri Valpola.
\newblock Mean teachers are better role models: Weight-averaged consistency
  targets improve semi-supervised deep learning results.
\newblock In {\em Advances in neural information processing systems}, pages
  1195--1204, 2017.

\bibitem{venkateswara2017Deep}
Hemanth Venkateswara, Jose Eusebio, Shayok Chakraborty, and Sethuraman
  Panchanathan.
\newblock Deep hashing network for unsupervised domain adaptation.
\newblock In {\em ({IEEE}) Conference on Computer Vision and Pattern
  Recognition ({CVPR})}, 2017.

\bibitem{wang2021tent}
Dequan Wang, Evan Shelhamer, Shaoteng Liu, Bruno Olshausen, and Trevor Darrell.
\newblock Tent: Fully test-time adaptation by entropy minimization.
\newblock In {\em International Conference on Learning Representations}, 2021.

\bibitem{xiao2012semi}
Min Xiao and Yuhong Guo.
\newblock Semi-supervised kernel matching for domain adaptation.
\newblock In {\em Twenty-Sixth AAAI Conference on Artificial Intelligence},
  2012.

\bibitem{Xu_Zhang_Ni_Li_Wang_Tian_Zhang_2020}
Minghao Xu, Jian Zhang, Bingbing Ni, Teng Li, Chengjie Wang, Qi Tian, and
  Wenjun Zhang.
\newblock Adversarial domain adaptation with domain mixup.
\newblock {\em Proceedings of the AAAI Conference on Artificial Intelligence},
  34(04):6502--6509, Apr. 2020.

\bibitem{zhang2018mixup}
Hongyi Zhang, Moustapha Cisse, Yann~N. Dauphin, and David Lopez-Paz.
\newblock mixup: Beyond empirical risk minimization.
\newblock In {\em International Conference on Learning Representations}, 2018.

\bibitem{zhang2021how}
Linjun Zhang, Zhun Deng, Kenji Kawaguchi, Amirata Ghorbani, and James Zou.
\newblock How does mixup help with robustness and generalization?
\newblock In {\em International Conference on Learning Representations}, 2021.

\bibitem{UEDA}
Yi~Yang Zhedong~Zheng.
\newblock Rectifying pseudo label learning via uncertainty estimation for
  domain adaptive semantic segmentation.
\newblock {\em International Journal of Computer Vision}, 2021.

\bibitem{zhuang2020comprehensive}
Fuzhen Zhuang, Zhiyuan Qi, Keyu Duan, Dongbo Xi, Yongchun Zhu, Hengshu Zhu, Hui
  Xiong, and Qing He.
\newblock A comprehensive survey on transfer learning, 2020.

\end{thebibliography}
}

\end{document}